\documentclass[journal,twoside,print]{ieeecolor}
\usepackage{arxiv}
\usepackage{cite}
\usepackage{amsmath,amssymb,amsfonts}
\usepackage{pifont}
\newcommand{\cmark}{\ding{51}}%
\newcommand{\xmark}{\ding{55}}%
\usepackage{setspace}
\usepackage{bbm}

\usepackage{algpseudocode,algorithm}
\MakeRobust{\Call}
\usepackage{graphicx}
\usepackage[font=small]{caption}
\usepackage{subcaption}
\usepackage{textcomp}
\usepackage{multirow}
\usepackage{colortbl}
\usepackage{hhline}
\usepackage{placeins}
\usepackage[table,xcdraw]{xcolor}

\def\BibTeX{{\rm B\kern-.05em{\sc i\kern-.025em b}\kern-.08em
    T\kern-.1667em\lower.7ex\hbox{E}\kern-.125emX}}
\begin{document}
\bstctlcite{IEEEexample:BSTcontrol}
\title{Latent Graph Representations for Critical View of Safety Assessment}
\author{Aditya Murali, Deepak Alapatt, Pietro Mascagni, Armine Vardazaryan, Alain Garcia, Nariaki Okamoto, Didier Mutter, and Nicolas Padoy
\thanks{This work was supported by French state funds managed by the ANR within the National AI Chair program under Grant ANR-20-CHIA-0029-01 (Chair AI4ORSafety) and within the Investments for the future program under Grants ANR-10-IDEX-0002-02 (IdEx Unistra) and ANR-10-IAHU-02 (IHU Strasbourg). This work was granted access to the HPC resources of IDRIS under the allocation 2021-AD011011640R1 made by GENCI.}
\thanks{Aditya Murali, Deepak Alapatt, and Nicolas Padoy are affiliated with ICube, University of Strasbourg, CNRS, France (email: \{murali, alapatt, npadoy\}@unistra.fr).}
\thanks{Pietro Mascagni, Armine Vardazaryan, Alain Garcia, Didier Mutter, and Nicolas Padoy are affiliated with IHU-Strasbourg, Institute of Image-Guided Surgery, Strasbourg, France.}
\thanks{Pietro Mascagni is affiliated with Fondazione Policlinico Universitario Agostino Gemelli IRCCS, Rome, Italy.}
\thanks{Nariaki Okamoto is affiliated with Institute for Research against Digestive Cancer (IRCAD), Strasbourg, France.} 
}

\maketitle

\begin{abstract}
Assessing the critical view of safety in laparoscopic cholecystectomy requires accurate identification and localization of key anatomical structures, reasoning about their geometric relationships to one another, and determining the quality of their exposure. 
Prior works have approached this task by including semantic segmentation as an intermediate step, using predicted segmentation masks to then predict the CVS.
While these methods are effective, they rely on extremely expensive ground-truth segmentation annotations and tend to fail when the predicted segmentation is incorrect, limiting generalization.
In this work, we propose a method for CVS prediction wherein we first represent a surgical image using a disentangled latent scene graph, then process this representation using a graph neural network.
Our graph representations explicitly encode semantic information -- object location, class information, geometric relations -- to improve anatomy-driven reasoning, as well as visual features to retain differentiability and thereby provide robustness to semantic errors.
Finally, to address annotation cost, we propose to train our method using only bounding box annotations, incorporating an auxiliary image reconstruction objective to learn fine-grained object boundaries.
We show that our method not only outperforms several baseline methods when trained with bounding box annotations, but also scales effectively when trained with segmentation masks, maintaining state-of-the-art performance.
\end{abstract}


\section{Introduction}
\label{sec:introduction}
Surgical video analysis is a quickly expanding research direction with several promising applications such as automated surgical phase, gesture, and tool recognition/segmentation~\cite{FunkeBOBWS19, jin2017sv, MaierHein2017, jin2019incorporating}.
One key next step is to develop video-based methods for safety-critical applications. A few such applications have been proposed, including detection of critical or adverse events~\cite{wei2021intraoperative, yu2022live} and automated identification of safe actions, clinical criteria, or anatomical regions~\cite{mascagni2021artificial,madani2022artificial, kolbinger2022artificial}. 
A common thread among these applications is the increased importance of anatomy, and often times fine-grained anatomy. Go-NoGo Net~\cite{madani2022artificial}, for instance, identifies regions of safe and unsafe dissection that are defined independently of surgical tool information, while Kolbinger et al.~\cite{kolbinger2022artificial} explore automatic identification of dissection planes, which are similarly defined based on anatomy rather than surgical tools or surgeon activity.
Yet, distinguishing anatomical structures is a difficult task as they are similar in color and texture and are highly deformable.
These approaches tackle this task using spatially dense annotations (e.g. segmentation masks), but such annotations require clinical expertise and are therefore prohibitively difficult to collect and review.

\begin{figure*}
\centering
\includegraphics[width=\textwidth]{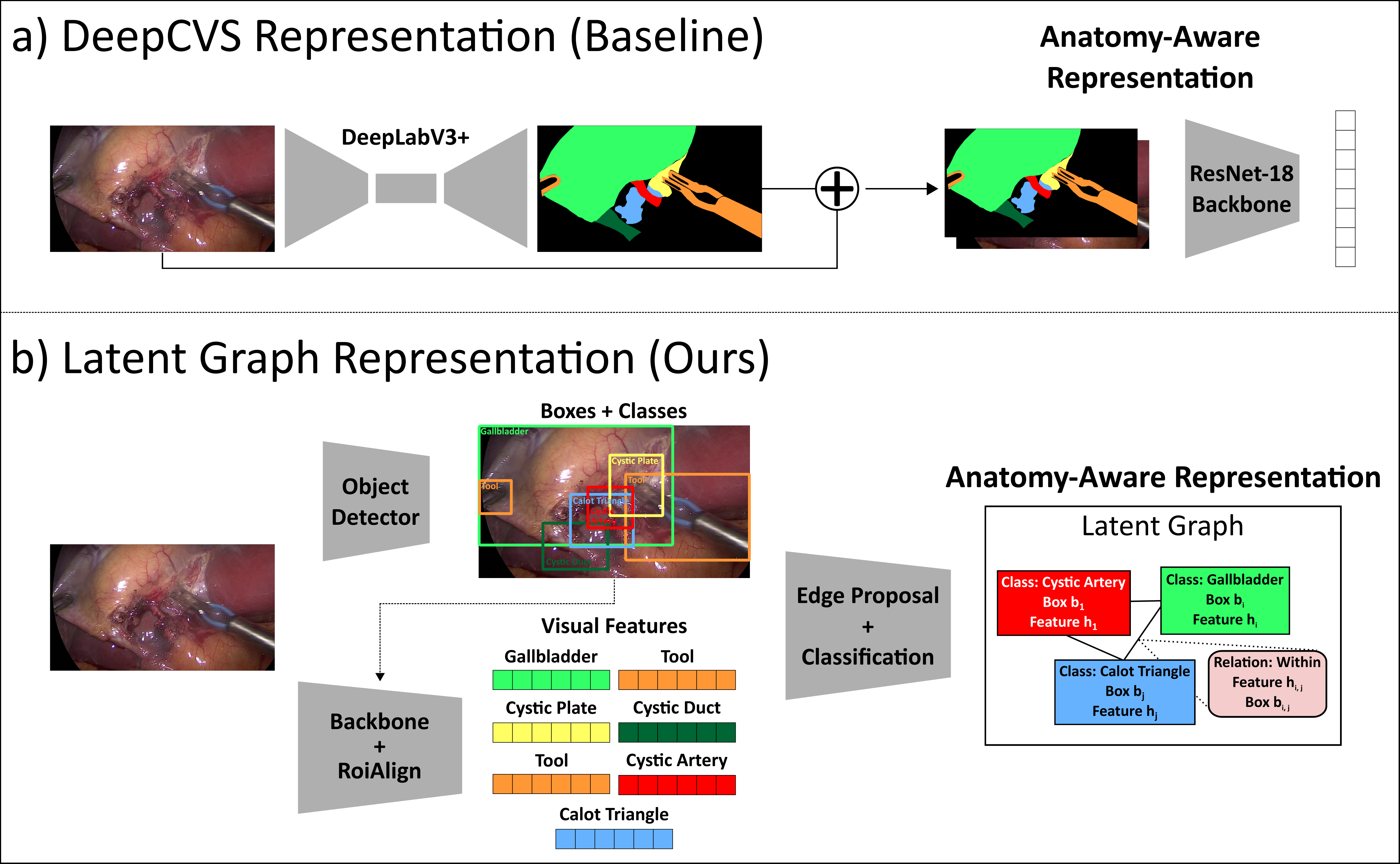}
\caption{Top: An illustration of the representation used in DeepCVS~\cite{mascagni2021artificial}, which consists of an image concatenated with the corresponding predicted segmentation mask (See Sec. \ref{subsec:baseline_models}). Bottom: Our proposed latent graphical representation that represents the surgical scene based on the anatomical structures and tools as well as their bounding box locations and visual features.}
\label{fig:baseline_vs_graph}
\end{figure*}

In this work, we aim to enable anatomy-driven reasoning without the need for expensive segmentation annotations, focusing on automated critical view of safety (CVS) assessment as our downstream task.
The CVS is a clinically validated intervention that is associated with reduced rates of bile duct injury in laparoscopic cholecystectomy; consequently, automatic assessment of CVS has been gaining attention in the surgical data science community. The recently proposed DeepCVS~\cite{mascagni2021artificial} approaches CVS prediction by extending the traditional classification paradigm, first concatenating an image with its segmentation mask before processing it with a shallow neural network to predict the CVS criteria.
To reduce annotation costs, the authors propose to annotate only a subset of images with dense segmentation masks, which they bootstrap to generate pseudo-labels for the remainder of images.
We take inspiration from this annotation approach and take it one step further, replacing the expensive segmentation masks with bounding box annotations to drastically reduce labeling costs.

Moving to bounding box annotations poses significant information loss that exacerbates a key limitation of DeepCVS: concatenating an image with its segmentation mask is an inefficient way of fusing visual and semantic information, as the model must implicitly learn to associate objects with their visual properties (see Fig. \ref{fig:baseline_vs_graph}a).
To tackle this issue, inspired by recent works on object-centric representations~\cite{wang2018videos,raboh2020differentiable,khan2021spatiotemporal}, we propose to construct an anatomy-aware \textit{latent graph representation} by running an object detector then using the predicted semantics (bounding boxes, class probabilities) to disentangle and re-structure the latent space of an image as a graph.
The resulting graph's nodes contain the class, location, and \textit{differentiable} visual features of each anatomical structure, while the edges encode geometric and \textit{differentiable} visual properties of the inter-structure relationships (see Fig. \ref{fig:baseline_vs_graph}b).
We train the latent graphs using the subset of data with bounding box annotations, as well as an auxiliary reconstruction objective that helps learn fine-grained object boundaries that are missing in the bounding box annotations.
Finally, we finetune our graph representations on the entire dataset to predict CVS, using a graph neural network for downstream prediction.
Our proposed approach thus learns a representation that seamlessly integrates visual and semantic information, enabling strong performance even without segmentation masks.

Additionally, DeepCVS is trained and validated on a small dataset of hand-picked frames, raising questions about its generalizability to real surgical scenarios. To properly evaluate CVS assessment performance, we introduce a dataset with 5x as many images, segmentation masks, and CVS annotations as that of DeepCVS, with frames sampled at even intervals rather than hand-picked for annotation.
This requires methods to generalize to blur, occlusions caused by tools, smoke, and bleeding, all of which naturally occur during a procedure.

We evaluate our model in two settings: CVS criteria prediction using (1) bounding box annotations, our primary task of interest, and (2) segmentation masks, to ensure that our method scales to different ground-truth availability scenarios.
Our proposed LatentGraph-CVS (LG-CVS), dramatically improves CVS prediction performance: the best baseline approach achieves an mAP of 62.8 when trained with segmentation masks, while LG-CVS achieves an mAP of 63.6 when trained with bounding boxes alone. Moreover, LG-CVS scales with better annotations, improving to an mAP of 67.3 when trained with segmentation masks.
Lastly, we demonstrate that our auxiliary reconstruction objective provides method-agnostic improvements in the label-efficient bounding box setting.

Our contributions can be summarized as follows:
\begin{enumerate}
  \item We propose a novel object-centric approach for CVS prediction that first encodes a surgical scene into an anatomy-aware latent graph representation, containing visual and semantic properties of each constituent object and inter-object relation, then applies a graph neural network to predict the CVS criteria.
  \item We enable CVS assessment using bounding box annotations rather than expensive segmentation masks, obtaining state-of-the-art performance for CVS criteria prediction with vastly lower annotation cost.
  \item We show that an auxiliary reconstruction objective can improve performance and stability across methods when training with bounding boxes.
\end{enumerate}

\section{Related Work}

\subsection{Scene Graphs}
Scene graphs have been used in numerous fashions in the computer vision community.
Initial works focused on scene graph generation from images~\cite{yang2018graph,zellers2018neural,lu2016visual} as an extension of object detection.
Later works expanded these efforts to dynamic/spatio-temporal scene graph generation~\cite{Ji_2020_CVPR, Cong_2021_ICCV}, aggregating detected scene graphs across time, and 3D scene graph generation~\cite{Wu_2021_CVPR} augmenting the predicted scene graphs with properties such as object shapes and depth.
These efforts have been translated to the surgical domain in a series of works that focuses on scene graph prediction from a dataset of benchtop robotic surgery videos~\cite{islam2020learning, seenivasan2022global}, as well as a recent work predicting 4D (spatiotemporal, 3D) scene graphs describing operating room activity and workflow~\cite{ozsoy20224d}.
In parallel, numerous works have shown the value of scene graphs as input to various downstream tasks.
\cite{johnson2018image,farshad2021migs} describe approaches to generate images from scene graphs, first generating a scene layout from an input graph then using this layout to reconstruct the image.
\cite{dhamo2020semantic} extends this paradigm to image editing using scene graphs, augmenting the scene graphs with visual features extracted from the original image.

Other works build on this idea of intermediate scene graph representations, using them for various downstream tasks.
\cite{raboh2020differentiable} uses an intermediate graph representation for visual relationship detection.
\cite{Gu_2019_CVPR} and~\cite{Khandelwal_2021_ICCV} propose to improve scene graph prediction by including image reconstruction/segmentation from the intermediate graph as auxiliary objectives.
\cite{wang2018videos} and~\cite{khan2021spatiotemporal} propose to use intermediate spatio-temporal graph representations for fine-grained action recognition; ~\cite{materzynska2020something} augments these representations with explicit semantic information (e.g. box coordinates, class probabilities), demonstrating improved action recognition performance.

Such approaches have also been applied in the surgical domain:~\cite{sarikaya2020towards} constructs spatio-temporal graphs using surgical tool pose information to condition downstream surgical activity recognition on tool trajectories, but the resulting graphs discard the visual information in the scene; meanwhile,~\cite{long2021relational} constructs multi-modal graphs using video and kinematic information for downstream gesture recognition, but these graphs are coarse-grained and do not include object information, instead representing each input modality as a node. \cite{pang2022rethinking} proposes to build a single-frame graph representation using class activation mapping to extract object-centric features; however, when trying to capture anatomical structures that co-occur in a majority of frames, such an approach is unlikely to generalize.

In this work, we translate these ideas and address their limitations, focusing on a new task: CVS prediction.
Unlike tasks like action recognition, which is explored in some of the aforementioned works, CVS prediction requires reasoning about the geometric configuration of anatomical structures in each frame; consequently, we encode these relationships in the latent graph edges rather than foregoing edges as in~\cite{materzynska2020something} or using them to encode temporal relationships as in~\cite{wang2018videos,khan2021spatiotemporal}.
In addition, because we use only bounding box annotations to train our graph representations, we include an auxiliary reconstruction objective to indirectly encode object boundary information in the node and edge visual features.
In our experiments, we conduct comprehensive ablation studies that illustrate the impact of these additions.

\subsection{Machine Learning for Surgical Safety Applications}
Surgical workflow recognition from video is well-explored through tasks such as phase recognition, step detection, tool detection/segmentation, and gesture recognition~\cite{jin2017sv, zisimopoulos2018deepphase, czempiel2020tecno, ramesh2021multi, FunkeBOBWS19}, with several works showing effective performance even in weakly supervised and unsupervised settings~\cite{liu2020unsupervised, sestini2022fun, wu2021cross, shi2021semi, ramesh2022dissecting}.
One missing component in these tasks is the explicit consideration of anatomy; to this end, several tasks have been introduced, including surgical triplet recognition, which involves decomposing surgical workflow into triplets of \textit{$\langle$instrument, verb, target$\rangle $}~\cite{nwoye2022cholectriplet2021}, and full scene segmentation~\cite{pfeiffer2019generating, hong2020cholecseg8k, grammatikopoulou2021cadis, alapatt2021temporally}.

Recent works have started to translate these ideas to safety-critical applications, which largely rely on effective anatomy recognition.
The aforementioned~\cite{madani2022artificial} trains a neural network to identify safe and unsafe dissection zones (Go-NoGo zones) in laparascopic cholecystectomy, while another recent work proposes a model for binary segmentation of critical structures (cystic artery and duct)~\cite{owen2021detection}, also in cholecystectomy.
While these approaches are effective, their outputs are not yet clinically validated to correlate to safer outcomes; in contrast, the critical view of safety, first proposed in 1995~\cite{strasberg1995analysis}, has a long history of effectiveness, and is now included as a component of all major guidelines for safe laparascopic cholecystectomy.
To this end, Mascagni et al. proposed EndoDigest~\cite{mascagni2021computer}, an approach to automatically isolate the period of cystic duct division in cholecystectomy procedures followed by the aforementioned DeepCVS~\cite{mascagni2021artificial} for CVS criteria prediction.

Our work falls into the umbrella of surgical video analysis for anatomy-based safety-critical applications, for which we take CVS prediction as a representative task. We present an improved methodology to incorporate anatomical information into surgical scene representations by constructing an anatomy-aware latent graph representation, which we can process with a graph neural network for downstream task (CVS) prediction.

\section{Methods}
\label{sec:methods}

In this section, we begin by describing our dataset for CVS criteria prediction.
Then we describe our latent graph encoder $\Phi_{\text{LG}}$ that generates a latent graph $G$ from an image $I$ (illustrated in Fig. \ref{fig:baseline_vs_graph}b).
Finally, we describe our two latent graph decoders: a reconstruction decoder $\phi_{\mathcal{R}}$ to help train the latent graphs and a CVS decoder $\phi_{\text{CVS}}$ to predict CVS (illustrated in Fig. \ref{fig:lg_decoders}).

\subsection{Dataset}
\label{subsec:dataset}

DeepCVS~\cite{mascagni2021artificial}, the state-of-the-art method for CVS prediction, utilizes a dataset of 2854 images that are hand-picked from 201 laparascopic cholecystectomy videos.
These images are annotated with achievement of each of the three CVS criteria, and a subset of 402 images are further annotated with semantic segmentation masks.
As previously noted, this is not only a small dataset but also an unrealistic representation of a surgical procedure, as the frame selection process introduces bias.
Alapatt et al.~\cite{alapatt2021temporally} recognize this problem and in a follow-up work, introduce the Endoscapes dataset, which comprises 1933 frames selected at a regular interval (once every 30 seconds) from the same 201 videos and annotated with semantic segmentation masks.
Because we are interested in CVS criteria prediction rather than semantic segmentation, we extend the Endoscapes dataset by annotating one frame every 5 seconds with CVS labels, which are vectors of three binary values corresponding to the three CVS criteria (C1, C2, C3 in Table \ref{table:dataset_stats}), and annotated by three independent and specifically trained surgeons following the protocol of~\cite{mascagni2021surgical}.

Then, to train our latent graph representations, we additionally generate ground-truth bounding box and scene graph annotations (see Fig. \ref{fig:qual}) from the segmentation masks. 
To generate bounding boxes, we take the semantic mask for each class and compute the smallest rectangular box that encloses the mask.
For classes with multiple instances (e.g. tools), we include a connected components step which separates each instance\footnote{In certain cases, the connected components algorithm fails to separate multiple instances because the masks of the instances overlap. In these cases, we annotate a single box for the overlapping instances.}.
Finally, to generate scene graphs, inspired by~\cite{johnson2018image}, we assign one of three directional relationships between each pair of boxes based on their bounding box coordinates: $\{\text{left-right}, \text{up-down}, \text{inside-outside}\}$.
Unlike~\cite{johnson2018image} we use undirected edges rather than directed and select $E$ edges per node by ranking edges using generalized intersection-over-union (gIoU)~\cite{Rezatofighi_2018_CVPR} between the two node bounding boxes; this ensures more consistent graphs, facilitating downstream CVS prediction.

Altogether, we obtain the \textit{Endoscapes+} dataset that is roughly 5x the scale of the dataset used in DeepCVS, comprising 11090 images with CVS annotations of which 1933 also have segmentation (from Endoscapes~\cite{alapatt2021temporally}), bounding box, and synthetic scene graph annotations.
For training and evaluation, we aggregate the three CVS annotations per image into a single ground-truth consensus vector by computing the mode for each criterion.
Finally, we split the 201 videos into 120 training, 41 validation, and 40 testing, adopting the same split as~\cite{alapatt2021temporally}.
Table \ref{table:dataset_stats} shows the proportion of frames in which each CVS criterion is achieved.

\begin{table}
\caption{Achievement Rates (\%) of each CVS Criterion.}
\label{table:dataset_stats}
\centering
\begin{tabular}{cccc} 
\textbf{Criterion} & \textbf{Train} & \textbf{Val} & \textbf{Test}  \\ 
\hline
C1: Two Structures       & 15.6           & 16.3         & 24.0           \\ 
C2: HCT Dissection       & 11.2           & 12.5         & 17.1           \\ 
C3: Cystic Plate         & 17.9           & 16.7         & 27.1           \\ 
\end{tabular}
\end{table}

\subsection{Latent Graph Encoder}
\label{subsec:latent_graph_encoder}

To obtain the latent graph representation $G$ of an image $I$, we pass the image through our latent graph encoder $\Phi_{\text{LG}}$.
$\Phi_{\text{LG}}$ consists of an object detector (augmented with a mask head when training with segmentation ground truth) to generate graph nodes, an edge proposal module to predict the graph structure, and finally a graph neural network that updates node and edge features and classifies each edge.
Running the object detector yields a set of $N$ objects\footnote{In cases where fewer than $N$ objects are detected, we set $N = \hat{N}$ where $\hat{N}$ is the number of detected objects.} or nodes $\mathcal{O} = \{b_i, c_i, m_i \, | \, 1 \leq i \leq N\}$, where $b_i \in \mathbb{R}^4$ are box coordinates, $c_i \in [0, 1]^C$ are vectors of class probabilities with $C$ object classes (including background), and $m_i \in \mathbb{R}^{M\times M}$ are instance masks.
We then augment these detected objects with visual features, following two separate strategies depending on the object detector used: for two-stage detectors (e.g. FasterRCNN~\cite{NIPS2015_14bfa6bb}), we apply the $RoIAlign$ operation~\cite{He_2017_ICCV} using the feature map of the image $H$ and the detected bounding boxes $b_i$ to yield node features $h_i \in \mathbb{R}^{\mathcal{F}_{\text{backbone}}}$; meanwhile, for transformer-based detectors (e.g. DETR~\cite{carion2020end}), we directly use the object queries from the transformer decoder to obtain the $h_i$. $\mathcal{F}_{\text{backbone}}$ here represents the dimension of the node features, which is also the output dimension of the backbone network (two-stage detectors) or the dimension of the object queries (transformer-based detectors).
Combining the $h_i$ with the $b_i$ and $c_i$ from the set of objects $\mathcal{O}$ yields the initial graph nodes $\mathcal{N}^{\text{init}}$:
\begin{equation}
    \label{equation:node_defn}
    \mathcal{N}^{\text{init}} = \{b_i, c_i, h_i \, | \, 1 \leq k \leq N\}.
\end{equation}

\noindent \textbf{Edge Proposal.} Once we have the graph nodes, the next step is to generate the graph structure by predicting the connectivity among nodes.
To do so, we start with a fully connected graph, then score each edge using a learned function $\phi_{\text{edge\_score}}$, then sample the highest scoring edges.
We implement $\phi_{\text{edge\_score}}$ as a Relation Proposal Network (RelPN)~\cite{yang2018graph} that takes as input the predicted bounding box, class logits, and visual features for a pair of nodes $i, j$ and outputs a scalar score $s_{i, j}$ for each edge $(i, j)$:
\begin{equation}
    \label{equation:edge_score}
    \begin{gathered}
        \phi_{\text{edge\_score}} : \mathbb{R}^{\mathcal{F}_\text{backbone} + C + 4} \times \mathbb{R}^{\mathcal{F}_\text{backbone} + C + 4} \mapsto \mathbb{R}^{1}, \\
        x_i = (b_i, c_i, h_i), \  x_j = (b_j, c_j, h_j), \\
        s_{i, j} = \phi_{\text{edge\_score}}(x_i, x_j).
    \end{gathered}
\end{equation}

We then select the $E$ highest scoring edges per node and drop duplicate edges (as $G$ is undirected, this sampling process can generate duplicates), to yield edge indices $\mathcal{E}_I \subseteq [1, N] \times [1, N]$.
Finally, for each edge, defined by the indices in $\mathcal{E}_I$, we compute a bounding box $b_{i, j}$, edge mask $\hat{m}_{i, j}$, and visual features $v_{i, j}$, yielding the initial graph edges $\mathcal{E}^{\text{init}}$:
\begin{equation}
    \label{equation:edge_proposal}
    \begin{gathered}
        b_{i, j} = b_{i} \cup b_{j}, \hat{m}_{i, j} = [\hat{m}_i; \hat{m}_j] \\
        h_{i, j} = \phi_{\text{edge\_feat\_extract}}(H, b_{i, j}, h_i, h_j), \\
        \mathcal{E}^{\text{init}} = \{e_{i, j}: [b_{i, j}; h_{i, j}] \, | \, (i, j) \in \mathcal{E}_I\},
    \end{gathered}
\end{equation}
where $\phi_{\text{edge\_feat\_extract}}$ either applies the $RoIAlign$ operation (two-stage detectors) or computes $h_{i, j}$ as the sum of $h_i$ and $h_j$ (transformer-based detectors).

To train $\phi_{\text{edge\_score}}$, we assign ground-truth binary edge presence values $\ell_{\text{presence}}(e_{i, j})$ to each edge $e_{i, j}$ based on the gIoU between its bounding box $b_{i, j}$ and the closest ground truth edge bounding box:
\begin{equation}
    \ell_{\text{presence}}(e_{i, j}) =
    \begin{cases}
        1& \text{if } \displaystyle \max_{k}{gIoU(b_{i, j}, b_k^{\text{gt}})} \geq 0.5,\\
        0& \text{otherwise}.
    \end{cases}
\end{equation}

\begin{figure*}
\centering
\includegraphics[width=\textwidth]{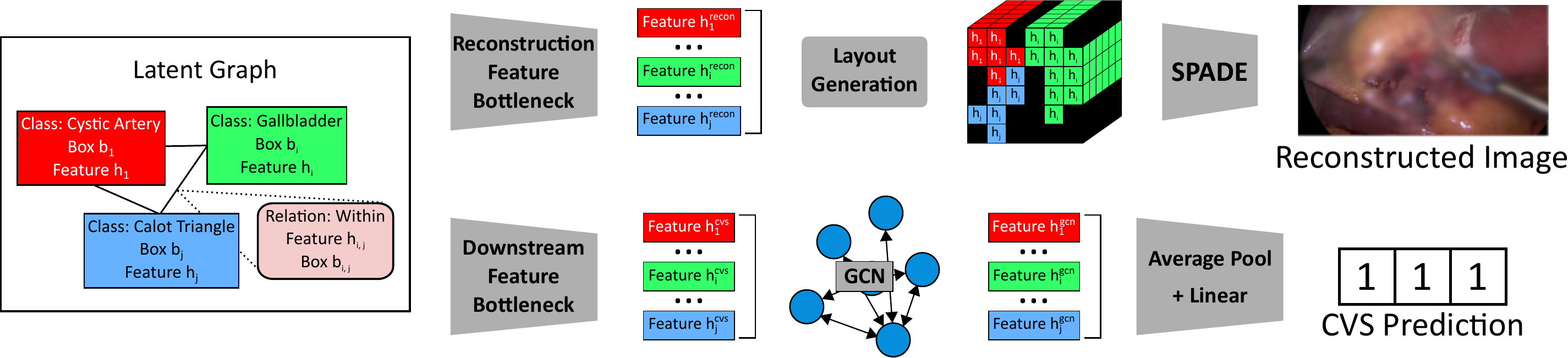}
\caption{The scene graph decoders $\phi_{\mathcal{R}}$ and $\phi_{\text{CVS}}$ to reconstruct an image and predict the CVS using the latent graph $G$.}
\label{fig:lg_decoders}
\end{figure*}

\noindent \textbf{Edge Classification.} Once we have our initial graph (comprising $\mathcal{N}^{\text{init}}$ and $\mathcal{E}^{\text{init}}$), we apply a 2-layer GNN, $\phi_{\text{LG-GNN}}$, to update the features of each node and edge based on the graph structure. This step is critical for effective edge classification, as the union box $b_{i, j}$ and corresponding visual feature $h_{i, j}^{\text{SG}}$ are not always sufficient to properly assess the relation between nodes $i$ and $j$.
We adopt the GNN architecture of~\cite{dhamo2020semantic}, which iteratively updates edge features and node features at each layer, and define the output dimension of $\phi_{\text{LG-GNN}}$ to be $\mathcal{F}_{\text{backbone}}$.
Applying $\phi_{\text{LG-GNN}}$ thus yields new node and edge features as follows:
\begin{equation}
    \label{equation:lg_gnn}
    \begin{gathered}
        H^{\text{LG}}_{\text{node}}, H^{\text{LG}}_{\text{edge}} = \phi_{\text{LG-GNN}}(\mathcal{N}^{\text{init}}, \mathcal{E}^{\text{init}}), \ \text{where} \\
        H^{\text{LG}}_{\text{node}} = \{h^{\text{LG}}_i \, | \, i \in [1, N]\},\  
        H^{\text{LG}}_{\text{edge}} = \{h^{\text{LG}}_{i, j} \, | \, (i, j) \in \mathcal{E}_I\}.
    \end{gathered}
\end{equation}

\noindent We pass the updated edge features through $\phi_{\text{edge\_classifier}}$, a 2-layer multi-layer perceptron to obtain edge class logits:
\begin{equation}
    \label{equation:edge_classification}
    \begin{gathered}
        \phi_{\text{edge\_classifier}}: \mathbb{R}^{\mathcal{F}_\text{backbone}} \mapsto \mathbb{R}^{C_E}, \\
        c_{i, j} = \phi_{\text{edge\_classifier}}(h^{\text{LG}}_{i, j}),
    \end{gathered}
\end{equation}
where $C_E$ is the number of relation classes.

To train $\phi_{\text{edge\_classifier}}$, we compute class labels for each edge following the same process as for $\ell_{\text{presence}}$, but assign the ground truth class as the target for positive matches:
\begin{equation}
    \ell_{\text{class}}(e_{i, j}) =
    \begin{cases}
        c_k^{\text{gt}}& \text{if } \displaystyle \max_{k}{gIoU(b_{i, j}, b_k^{\text{gt}})} \geq 0.5,\\
        0& \text{otherwise}.
    \end{cases}
\end{equation}

\noindent \textbf{Final Latent Graph Representation.} Our final latent graph representation $G$ is composed of the bounding boxes, class logits, and segmentation-grounded and GNN-processed visual features for each node and edge:
\begin{equation}
    \label{equation:final_latent_graph}
    \begin{gathered}
        G = \{\mathcal{N}^{\text{LG}}, \mathcal{E}^{\text{LG}}\},\ \text{where} \\
        \mathcal{N}^{\text{LG}} = \{(b_i, c_i, h^{\text{LG}}_i \, | \, i \in [1, N]\}, \text{and}\\
        \mathcal{E}^{\text{LG}} = \{(b_{i, j}, c_{i, j}, h^{\text{LG}}_{i, j} \, | \, (i, j) \in \mathcal{E}_I\}.
    \end{gathered}
\end{equation}

Figure \ref{fig:baseline_vs_graph}b summarizes the process of generating a latent graph $G$ from image $I$.

\subsection{Latent Graph Decoders}
\label{subsec:graph_decoders}

We introduce two separate decoders that take the latent graph $G$ as input: an image reconstruction decoder $\phi_{\mathcal{R}}$ and a CVS decoder $\phi_{\text{CVS}}$ (see Fig. \ref{fig:lg_decoders}).

\noindent \textbf{Image Reconstruction.} When training with bounding box labels alone, information about object boundaries is absent in the ground truth and therefore not encoded in the learned latent graph representation $G$.
This is particularly problematic for downstream CVS prediction, which relies on distinguishing fine-grained anatomical structures such as the hepatocystic triangle and cystic plate.
To address this problem, we introduce image reconstruction as an auxiliary task, generating an image $\hat{I}$ from the latent graph $G$, an image layout $L$, and a backgroundized image $I_\text{bg}$ to focus on the objects in the scene.

We follow the method of~\cite{dhamo2020semantic}, starting by computing $L$ using the predicted bounding boxes or instance masks.
Then, we use $L$ to spatially arrange the node features into a pixel-wise feature layout, $L_\text{feat}$, which we finally pass to a SPADE~\cite{park2019SPADE} image reconstruction module $\phi_{\mathcal{R}}$ (see Fig. \ref{fig:lg_decoders}) along with a backgroundized image $I_\text{bg}$ to reconstruct the image $I$.
We compute the image layout $L$, which stores all nodes\footnote{When we are using predicted bounding boxes to generate $L$, we can have overlapping boxes, hence the need to account for multiple indices in some pixels.} associated with a pixel $(p_x, p_y)$, as follows:
\begin{equation}
    \label{equation:layout_box}
    \begin{gathered}
        (x_1^i, y_1^i, x_2^i, y_2^i) = b_i\\
        L[p_x, p_y, i] =
        \begin{cases}
            1& \text{if }
                \begin{aligned}[t]
                    x_1^i &\leq p_x \leq x_2^i,\\
                    y_1^i &\leq p_y \leq y_2^i,
                \end{aligned}\\
            0& \text{otherwise}.
        \end{cases}
    \end{gathered}
\end{equation}

Using $L$, we can compute the feature layout $L_{\text{feat}}$ using the node visual features; here, we additionally apply a linear bottleneck $\mathcal{B}_{\mathcal{R}}: \mathbb{R}^{\mathcal{F}_\text{backbone}} \mapsto \mathbb{R}^{\mathcal{F}_\mathcal{R}}$ to the visual features to prevent the model from converging to a trivial solution. 
Altogether, we have:
\begin{equation}
    \begin{gathered}
        h^{\mathcal{R}}_i = \mathcal{B}_{\mathcal{R}}(h^{\text{LG}}_i), \\
        L_{\text{feat}}[p_x, p_y] = \displaystyle \sum_{i}{L[p_x, p_y, i] * h_i^{\mathcal{R}}}.
    \end{gathered}
\end{equation}
Finally, we compute the backgroundized image $I_{\text{bg}}$ by replacing all ground-truth bounding box regions with Gaussian noise; when ground-truth foreground is unavailable, we instead replace the predicted bounding box regions with noise.
Concatenating $L$, $L_{\text{feat}}$, and $I_{\text{bg}}$ and passing the result through $\phi_{\mathcal{R}}$ yields the reconstructed image $\hat{I}$.

\noindent \textbf{CVS Criteria Prediction.} To predict the CVS criteria from the latent graph, we use the CVS decoder $\phi_{\text{CVS}}$ (see Fig. \ref{fig:lg_decoders}), which is composed of a GNN (same architecture as $\phi_{\text{LG-GNN}}$) followed by a global pooling layer across node features and a linear layer that outputs a vector of three values, corresponding to predicted scores for each criterion.
Similarly to the image reconstruction decoding process, we apply feature bottlenecks $\mathcal{B}^{\mathcal{N}}_{\text{CVS}}: \mathbb{R}^{\mathcal{F}_\text{backbone}} \mapsto \mathbb{R}^{\mathcal{F}_\mathcal{N}}$, and $\mathcal{B}^{\mathcal{E}}_{\text{CVS}}: \mathbb{R}^{\mathcal{F}_\text{backbone}} \mapsto \mathbb{R}^{\mathcal{F}_\mathcal{E}}$ to the node and edge visual features before evaluating $\phi_{\text{CVS}}$.

\subsection{Training Process}
\label{subsec:training_process}

We train our model in 2 stages.
In the first stage, we train the latent graph encoder $\Phi_{\text{LG}}$ using the bounding box/segmentation and synthetic scene graph annotations.

Then, in the second stage, we freeze the object detector in the latent graph encoder and fine-tune the remaining components using the CVS labels and auxiliary reconstruction objective.
In order to enable end-to-end training while retaining object detection performance, we initialize a separate backbone (same architecture) for CVS prediction called $\phi_{\text{trainable\_backbone}}$.
We then freeze the original $\phi_{\text{backbone}}$ and use it along with the rest of the object detector components to detect objects, while using the features from $\phi_{\text{trainable\_backbone}}$ to extract the visual features used in the latent graph $G$. In practice, we also initialize the weights of $\phi_{\text{trainable\_backbone}}$ using the weights of $\phi_{\text{backbone}}$.

The overall loss function for the first stage is as follows:
\begin{equation}
    \label{equation:loss}
    \mathcal{L}_{\text{1}} = \mathcal{L}_{\text{object\_detector}} + \mathcal{L}_{\text{RelPN}} + \mathcal{L}_{\text{edge\_classifier}},
\end{equation}
where $\mathcal{L}_{\text{object\_detector}}$ is the loss formulation of the underlying object detector, $\mathcal{L}_{\text{RelPN}}$ is a binary cross entropy loss following~\cite{yang2018graph}, and $\mathcal{L}_{\text{edge\_classifier}}$ is the cross entropy loss between predicted relations and matched ground truth relations.

In the second stage, we freeze the weights of the object detector in $\Phi_{\text{LG}}$ and fine-tune the remaining weights as well as $\phi_{\text{trainable\_backbone}}$, the CVS decoder $\phi_{\text{CVS}}$ and reconstruction decoder $\phi_{\mathcal{R}}$.
We formulate the reconstruction loss $\mathcal{L}_{\text{reconstruction}}$ as the sum of the L1 Loss, Perceptual Loss~\cite{zhang2018perceptual}, and Structural Similarity loss~\cite{wang2004image} between the original and reconstructed images, and the CVS prediction loss $\mathcal{L}_{\text{CVS}}$ as a weighted binary cross entropy loss function, with loss weights given by inverse frequency balancing.

\textbf{Box Perturbation.} Our computed latent graph representations $G$ are sometimes characterized by semantic errors (missed objects, incorrectly localized objects, etc.), especially when the underlying object detector is less effective.
To help increase robustness to these errors, we randomly perturb the boxes $b_i \in \mathcal{N}_{\text{init}}$ predicted by the object detector during training.
Specifically, we add a random noise vector $p_i = \lambda_{\text{perturb}} * \left(p_{x}^{1}, p_{y}^{1}, p_{x}^{2}, p_{y}^{2}\right)$ to each box $b_i$, where the $p_x$ are drawn from a uniform distribution $U(-w, w)$, and the $p_y$ from $U(-h, h)$, where $h, w$ are the height and width of $b_i$.
We show the impact of this box perturbation on CVS criteria prediction and investigate the effect of the perturbation factor $\lambda_{\text{perturb}}$ in Section \ref{subsec:ablations}.

\section{Experiments and Results}
\label{sec:exps_and_results}

\FloatBarrier
\begin{table}
\centering
\caption{Effect of varying DeepCVS architecture (DeepLabV3+ Segmentation Model).}
\label{tab:deepcvs_mod}
\begin{tabular}{cc}
\textbf{Architecture} & \textbf{CVS Criteria mAP} \\ \hline
DeepCVS~\cite{mascagni2021artificial} & 54.2 \\
DeepCVS-MobileNetV3+ & 57.5 \\
DeepCVS-ResNet18 & \textbf{59.1} \\
DeepCVS-ResNet50 & 58.8 \\
\end{tabular}
\end{table}

\subsection{Baseline Models}
\label{subsec:baseline_models}

In this section, we describe a series of baseline methods, which are characterized by different methodologies of utilizing the dense labels (bounding box/segmentation) for CVS prediction.

\noindent \textbf{ResNet50 Classifier.} The most simple baseline is an ImageNet pre-trained ResNet50 classifier finetuned on our dataset for CVS prediction.
This method does not require any dense labels (bounding box or segmentation), making it the most label-efficient methodology.
The purpose of including this method is to illustrate the substantial improvements that can be gained by utilizing more fine-grained ground truth annotations.
We additionally include a variant of this model trained with an auxiliary reconstruction objective for consistency.

\noindent \textbf{DeepCVS.} DeepCVS~\cite{mascagni2021artificial}, the current state-of-the-art method for CVS criteria prediction, first passes images through a DeepLabV3+ model to predict segmentation masks, then concatenates the predicted masks with the original images and passes the result to a custom 6-layer network to predict the CVS criteria (see Fig. \ref{fig:baseline_vs_graph}a).
We note that different architectures may be better suited for this downstream classification task, and investigate replacing the original classification head from DeepCVS with MobileNetV3+, ResNet-18, and ResNet-50 architectures (see Table \ref{tab:deepcvs_mod}), modifying the initial layer to ingest a 10-channel input (image and mask with 7 channels).
We find that the adapted ResNet-18 architecture achieves the best performance on our dataset, and proceed to use this variant of DeepCVS for our main experiments.

\begin{algorithm}
\caption{Bounding Boxes to Layout}
\label{algorithm:box_mask}
\begin{algorithmic}
\State $L^{\text{box}} \gets $ \Call{zeros}{$H$, $W$, $C$}
\For {$b_i, c_i \in B, C$}
    \State $(x_1, y_1, x_2, y_2) \gets b_i$
    \State $L^{\text{box}}[y_1:y_2, x_1:x_2, c_i] \gets 1$
\EndFor
\end{algorithmic}
\end{algorithm}

Additionally, to enable fair comparisons with our method,
we replace the DeepLabV3+ segmentation model with Faster-RCNN (bounding box setting) or Mask-RCNN (segmentation setting) for the experiments in Table \ref{tab:main_exps}.
To extend DeepCVS to the bounding box setting, we convert the predicted bounding boxes into instance masks with the dimension of the input image then collapse the instance masks into a semantic layout as used in our reconstruction decoder (see Sec. \ref{subsec:graph_decoders}). 
Algorithm \ref{algorithm:box_mask} describes this process concretely, starting from the set of predicted bounding boxes $B$ and predicted classes $C$ for image $I$ and ending up with a scene layout $L$.
We follow the latter steps of the process in the segmentation setting (to handle instance masks instead of segmentation masks).

Lastly, we adapt our reconstruction objective to work with DeepCVS. We pass the aforementioned layout $L$ and the ResNet18 image features (use same feature for all objects) to $\phi_{\mathcal{R}}$ along with a backgroundized image $I_{\text{bg}}$, both as defined as in Sec. \ref{subsec:graph_decoders}.
We apply a linear bottleneck $\mathcal{B}^{\text{DeepCVS}}$ to the ResNet18 features, with the bottleneck size computed as $S = N * \mathcal{F}_{\mathcal{N}}$, ensuring that $S$ matches the total size of the latent graph $G$ in our method. Algorithm \ref{algorithm:deepcvs_reconstruction} describes this reconstruction formulation concretely.

\begin{algorithm}
\setstretch{1.35}
\caption{DeepCVS Reconstruction Decoder}
\label{algorithm:deepcvs_reconstruction}
\begin{algorithmic}
    \State $H^{\text{DeepCVS}} \gets \mathcal{B}_{\text{DeepCVS}}\Big(\phi_{\text{backbone}}^{\text{DeepCVS}}\big($\Call{concat}{$I, L$}$\big)\Big)$
    
    \State $L_{\text{feat}}^{\text{DeepCVS}} \gets$ \Call{zeros}{$S$, $H$, $W$}
    
    \State $L_{\text{feat}}^{\text{DeepCVS}}[:, L \neq 0] \gets H^{\text{DeepCVS}}$ \Comment{foreground $\leftarrow$ img feat}

    \State $I_{\text{bg}} \gets I$; $I_{\text{bg}}[:, L^{\text{box}} \neq 0] \gets 0$ \Comment{foreground $\leftarrow$ noise}
    
    \State $\hat{I} \gets \phi_{\mathcal{R}}\Big($\Call{concat}{$L, L_{\text{feat}}^{\text{DeepCVS}}, I_{\text{bg}}$}$\Big)$
\setstretch{1}
\end{algorithmic}
\end{algorithm}

\noindent 
\textbf{LayoutCVS.} We introduce an additional baseline which follows DeepCVS and the presented extensions exactly except uses only the layout $L$ as an input to the downstream ResNet18 model, bypassing concatenation with the original image.
We include this baseline to illustrate the importance of the visual features for CVS prediction and evaluate the effectiveness of DeepCVS in fusing semantic and visual information for downstream prediction.

\noindent 
\textbf{ResNet50-DetInit.} Lastly, we include a multi-task learning baseline, ResNet50-\textbf{Det}ector\textbf{Init}, which represents another approach to make use of bounding box/segmentation labels for CVS prediction.
This method is identical to the simple ResNet50 classifier described earlier but the ResNet50 weights are initialized from the backbone of the trained object detection/segmentation model.

\noindent We call our proposed approach, which uses a latent graph representation to predict CVS, \textbf{LG-CVS} (\textbf{L}atent\textbf{G}raph-CVS).

\begin{table}
\centering

\caption{CVS Criteria Prediction performance using (1) only CVS labels, (2) bounding box ground truth and (3) segmentation ground truth. Standard deviation shown across 3 randomly seeded runs.}
\label{tab:main_exps}
\resizebox{0.5\textwidth}{!}{
\begin{tabular}{cccc}
\multirow{2}{*}{Dense Labels} & \multirow{2}{*}{Method} & \multicolumn{2}{c}{CVS mAP} \\
 &  & \begin{tabular}[c]{@{}c@{}}No\\ Recon\end{tabular} & \begin{tabular}[c]{@{}c@{}}With\\ Recon\end{tabular} \\ \hline
None & ResNet50 & 51.7 $\pm$ 0.9 & 52.2 $\pm$ 0.3 \\ \hline
\multirow{4}{*}{\begin{tabular}[c]{@{}c@{}}Bounding\\ Boxes\end{tabular}} & LayoutCVS & 48.4 $\pm$ 0.4 & 49.6 $\pm$ 0.8 \\
 & DeepCVS & 51.2 $\pm$ 0.7 & 54.1 $\pm$ 1.3 \\
 & ResNet50-DetInit & 57.1 $\pm$ 1.3 & 57.6 $\pm$ 1.2 \\
 & \textbf{LG-CVS} & \textbf{60.7 $\pm$ 1.6} & \textbf{63.6 $\pm$ 0.8} \\ \hline
\multirow{4}{*}{\begin{tabular}[c]{@{}c@{}}Segmentation\\ Masks\end{tabular}} & LayoutCVS & 56.9 $\pm$ 0.6 & 56.1 $\pm$ 0.7 \\
 & DeepCVS & 60.0 $\pm$ 1.3 & 60.2 $\pm$ 1.6 \\
 & ResNet50-DetInit & 62.8 $\pm$ 1.1 & 60.5 $\pm$ 1.4 \\
 & \textbf{LG-CVS} & \textbf{67.7 $\pm$ 2.1} & \textbf{67.3 $\pm$ 1.4}
\end{tabular}}
\end{table}

\subsection{CVS Criteria Prediction}
\label{subsec:cvs_prediction}
We evaluate all methods in two experimental settings: using bounding box annotations vs. segmentation masks.
For each setting, we additionally present results with and without the auxiliary reconstruction objective. We use COCO mean average precision (mAP) to evaluate our various object detectors (bounding box and instance segmentation), Recall@10~\cite{lu2016visual} to evaluate scene graph prediction performance, and classification mAP to evaluate CVS prediction (average per-criterion average precision scores).
Table \ref{tab:main_exps} shows the results of each method with and without the reconstruction objective, using Faster-RCNN or Mask-RCNN as the underlying object detector for the bounding box and segmentation settings respectively.
Table \ref{tab:obj_detectors} extends this analysis to a variety of object detectors, showing the performance of all models with the reconstruction objective, as well as the performance of the underlying object detectors for the various tasks.
Last but not least, Table \ref{table:detailed_results} shows the performance of LG-CVS and DeepCVS (previous state-of-the-art) for each individual CVS criterion.

\noindent \textbf{Bounding Box Setting.} LG-CVS is particularly effective in the bounding box setting, achieving an improvement of \textbf{6.0} mAP over the best baseline, ResNet50-DetInit. Of note, both the LayoutCVS and DeepCVS are quite ineffective in this setting, with LayoutCVS performing worse than the simple ResNet50 baseline (which does not require bounding box annotations) and DeepCVS only outperforming the simple ResNet50 by 1.9 mAP. These results highlight the importance of visual features for CVS prediction and that DeepCVS fails to effectively fuse this visual information with the semantic information provided by the layout. The ResNet50-DetInit fares much better in this setting, but is still considerably worse than LG-CVS; this latter difference shows the impact of our graph-based modeling.

\noindent \textbf{Segmentation Setting.} LG-CVS can also make use of segmentation mask annotations when available, obtaining a performance boost of \textbf{4.1} mAP over LG-CVS trained with boxes, and \textbf{4.9} mAP over the best baseline method, ResNet50-DetInit. LayoutCVS and DeepCVS are much more effective when trained with segmentation masks, but still do not match the performance of the ResNet50-DetInit. This can be explained by differing benefits brought by their respective methodologies: LayoutCVS and DeepCVS are able to strongly condition CVS prediction on the underlying anatomy, which is very effective when the object detector accurately distinguishes the various anatomical structures, as in the segmentation setting; meanwhile, ResNet50-DetInit is able to leverage the effective visual representation learned by the object detector for CVS prediction. LG-CVS effectively combines the benefits of both classes of approach by encoding scene semantics as well as the visual features learned by the object detector in its latent graph. As a result, LG-CVS trained with bounding boxes alone outperforms all the baseline methods trained with segmentation masks.

\begin{table*}[t]
\centering

\caption{Extensions to Various Object Detectors. All results in the bounding box setting are shown with the reconstruction objective, while those in the segmentation setting do not use the reconstruction objective.}
\label{tab:obj_detectors}
\resizebox{\textwidth}{!}{
\begin{tabular}{ccccccccc}
\multirow{2}{*}{Dense Labels} & \multirow{2}{*}{Detector} & \multicolumn{2}{c}{\begin{tabular}[c]{@{}c@{}}Detection \\ (mAP)\end{tabular}} & \multirow{2}{*}{\begin{tabular}[c]{@{}c@{}}Semantic\\ Segmentation\\ (Dice)\end{tabular}} & \multirow{2}{*}{\begin{tabular}[c]{@{}c@{}}Scene Graph\\ Prediction\\ (Recall@10)\end{tabular}} & \multicolumn{3}{c}{CVS mAP} \\
 &  & Bbox & Segm &  &  & DeepCVS & \begin{tabular}[c]{@{}c@{}}ResNet50-DetInit\end{tabular} & \begin{tabular}[c]{@{}c@{}}LG-CVS\\ (Ours)\end{tabular} \\ \hline
\multirow{3}{*}{\begin{tabular}[c]{@{}c@{}}Bounding\\ Boxes\end{tabular}} & Faster-RCNN~\cite{NIPS2015_14bfa6bb} (Base) & 29.0 & \multirow{3}{*}{-} & \multirow{3}{*}{-} & 51.4 & 54.1 & 57.6 & \textbf{63.6} \\
 & Cascade-RCNN~\cite{cai2018cascade} & 30.4 &  &  & 51.7 & 53.2 & 57.0 & \textbf{64.3} \\
 & Deformable-DETR~\cite{zhu2020deformable} & 34.8 &  &  & 52.6 & 55.7 & 56.7 & \textbf{63.0} \\ \hline
\multirow{4}{*}{\begin{tabular}[c]{@{}c@{}}Segmentation\\ Masks\end{tabular}} & Mask-RCNN~\cite{He_2017_ICCV} (Base) & 30.9 & 32.5 & 71.50 & 51.5 & 60.9 & 60.5 & \textbf{67.3} \\
 & Cascade-Mask-RCNN~\cite{cai2018cascade} & 32.5 & 32.8 & 72.78 & 52.0 & 61.5 & 60.2 & \textbf{65.4} \\
 & Mask2Former~\cite{cheng2022masked} & 34.8 & 35.4 & 74.95 & 53.4 & 62.3 & 64.3 & \textbf{65.5} \\
 & DeepLabV3+~\cite{chen2018encoder} & \multicolumn{2}{c}{-} & 73.20 & N/A & 60.2 & 61.4 & -
\end{tabular}}
\end{table*}

\renewcommand{\arraystretch}{1.2}
\begin{table}
\centering
\caption{Detailed Comparison of DeepCVS and LG-CVS in Segmentation Setting (mAP is Mean Average Precision, Bacc is Balanced Accuracy, Mask-RCNN is detector for both).}
\label{table:detailed_results}
\begin{tabular}{ccccc}
\multicolumn{2}{c}{} & DeepCVS & ResNet50-DetInit & LG-CVS \\ \hline
\multirow{4}{*}{\rotatebox[origin=c]{90}{mAP}} & C1 & 65.9 & 57.3 & \textbf{69.5} \\
 & C2 & 52.6 & 54.9 & \textbf{60.7}\\
 & C3 & 61.8 & 69.3 & \textbf{71.8}\\
 & Avg & 60.2 & 60.5 & \textbf{67.3}\\\hline
\multirow{4}{*}{\begin{tabular}[c]{@{}c@{}}\rotatebox[origin=c]{90}{Bacc}\end{tabular}} & C1 & 74.0 & 69.0 & \textbf{78.6} \\
 & C2 & 73.4 & 70.2 & \textbf{81.4} \\
 & C3 & 70.7 & 74.3 & \textbf{79.4} \\
 & Avg & 72.3 & 71.2 & \textbf{79.8}
\end{tabular}
\end{table}
\renewcommand{\arraystretch}{1.0}

\noindent \textbf{Impact of Reconstruction Objective.} As shown in Table \ref{tab:main_exps}, the auxiliary reconstruction objective provides method-agnostic improvements in the bounding box setting, but is ineffective in the segmentation setting. This can be explained by the fact that we design the auxiliary reconstruction objective primarily for the bounding box setting, where it can help bridge the gap between the coarse bounding box annotations and fine-grained segmentation masks. In the segmentation setting, we already have the fine-grained object boundary information, and consequently we do not see an improvement.

\begin{figure*}
\centering
\includegraphics[width=\textwidth]{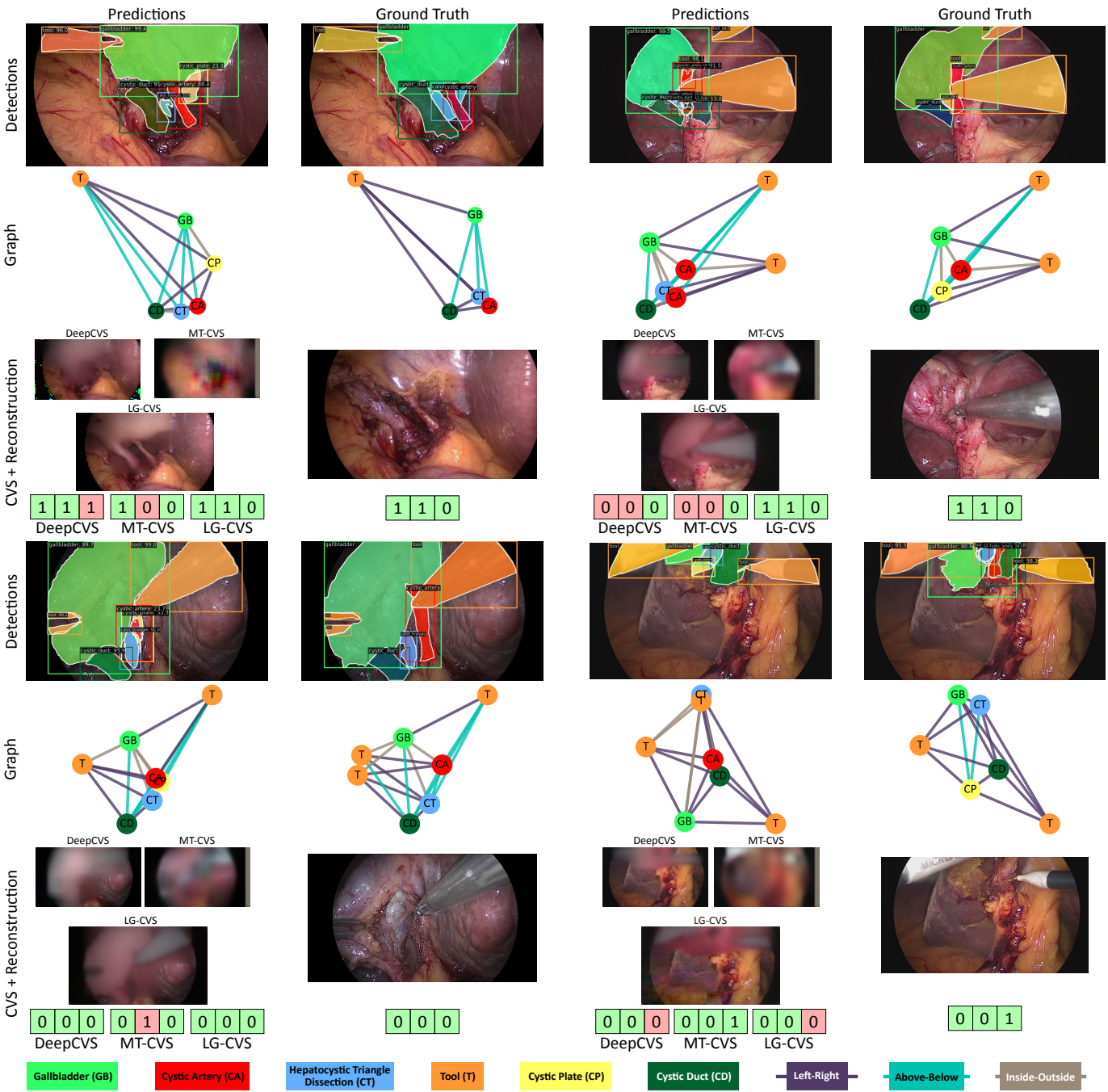}
\caption{Qualitative performance of DeepCVS, ResNet50-DetInit and LG-CVS (two best baselines and our method) using Mask-RCNN. Detections are shared across methods while predicted graphs are only used for LG-CVS. We also show the reconstructed images for each method. Edges are color coded with the relations described in the legend at the bottom. The CVS predictions represent C1, C2, C3 from left to right.}
\label{fig:qual}
\end{figure*}

Of note, the reconstructed images of ResNet50-DetInit and DeepCVS are considerably worse than those of LG-CVS, despite the fact that all the methods use the same bottlenecked representation size.
Figure \ref{fig:qual} qualitatively illustrates this phenomenon - the ResNet50-DetInit reconstructions are uniformly blurry, while the DeepCVS reconstructions lack detail especially in the anatomical structures; the LG-CVS reconstructions, on the other hand, more clearly delineate the anatomical structures (cystic artery, cystic duct, hepatocystic triangle).
This reinforces the notion that our disentangled latent graph representation more efficiently captures the anatomy of a scene (both visual and semantic properties) than a non-object centric feature map (DeepCVS and ResNet50-DetInit).

\begin{figure*}[t]
    \centering
    \begin{subfigure}[b]{0.32\textwidth}
        \centering
        \includegraphics[width=\textwidth]{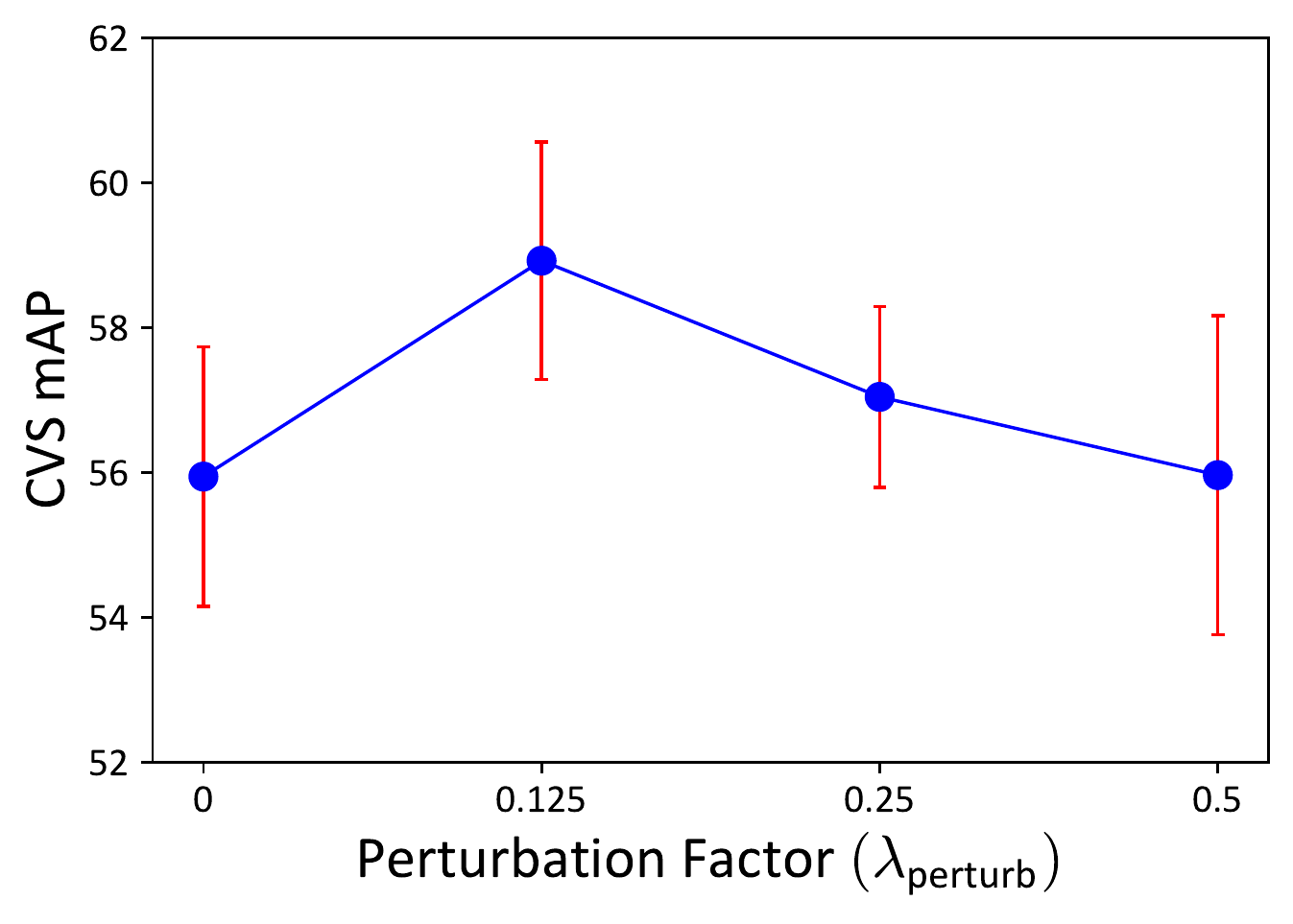}
        \caption{Effect of box perturbation factor $\lambda_{\text{perturb}}$ on downstream CVS performance.}
        \label{fig:box_perturbation}
    \end{subfigure}
    \begin{subfigure}[b]{0.32\textwidth}
        \centering
        \includegraphics[width=\textwidth]{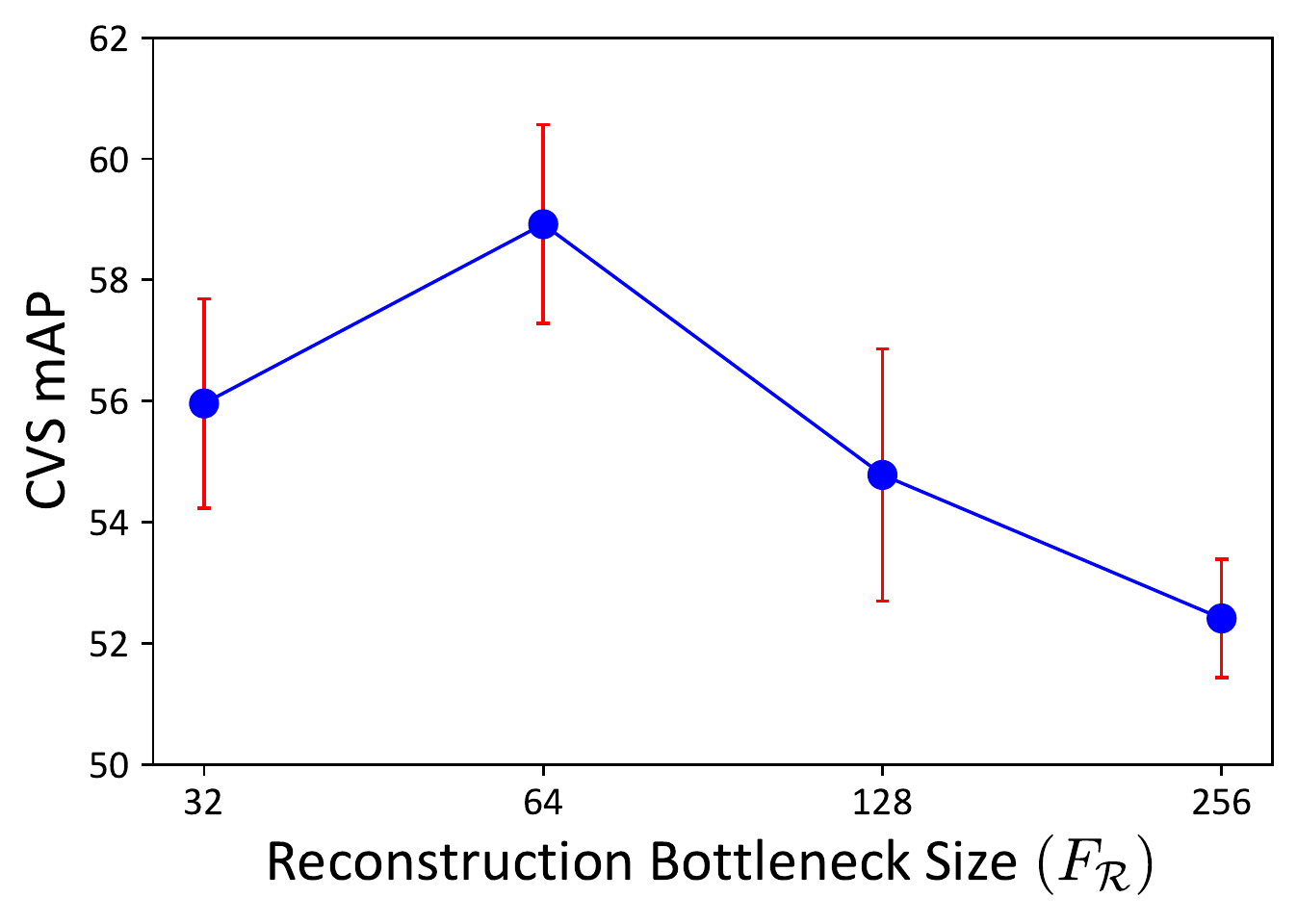}
        \caption{Effect of Reconstruction Bottleneck Size $\mathcal{F}_\mathcal{R}$ on downstream CVS performance.}
        \label{fig:reconstruction_bottleneck_size}
    \end{subfigure}
    \begin{subfigure}[b]{0.32\textwidth}
        \centering
        \includegraphics[width=\textwidth]{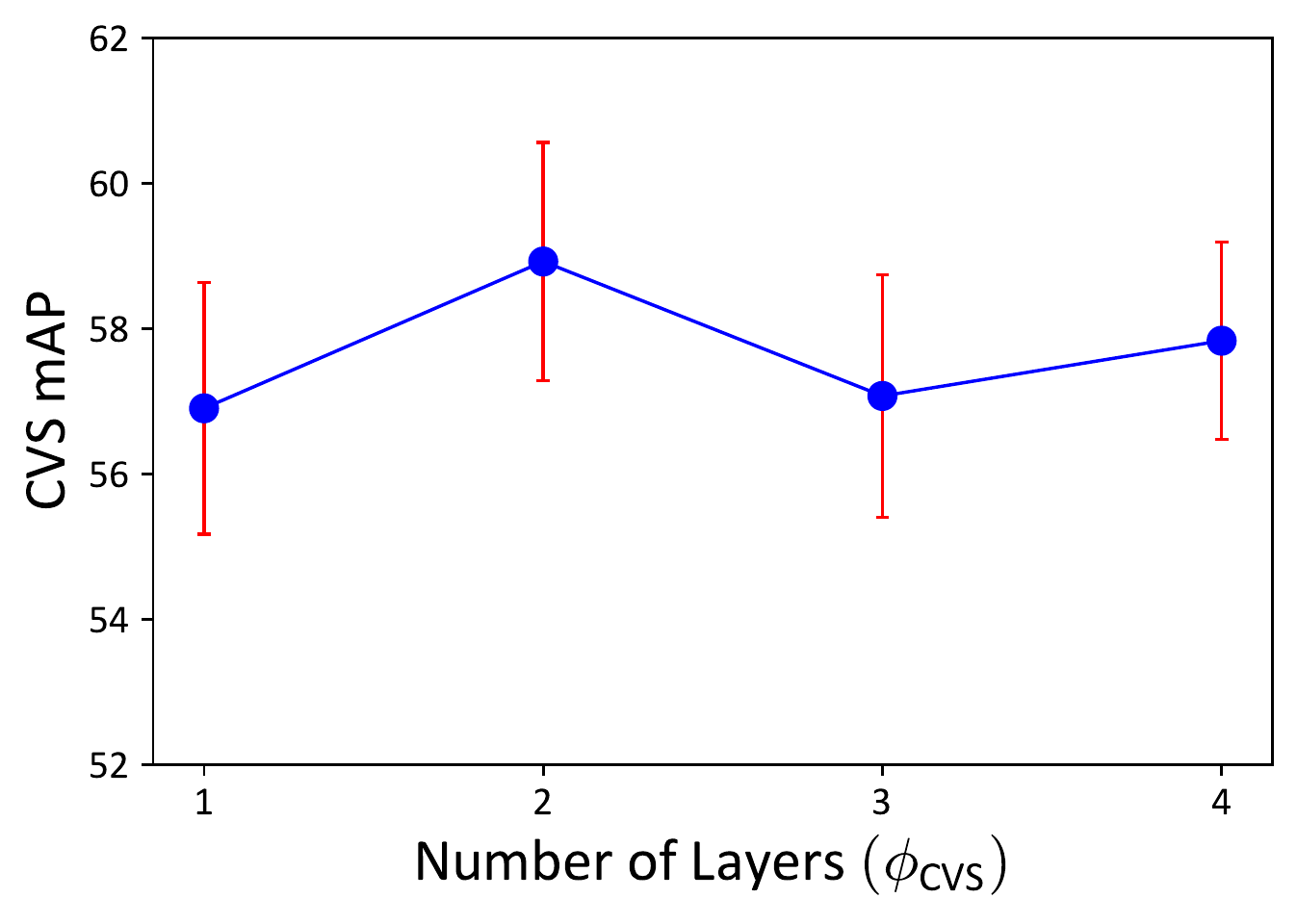}
        \caption{Effect of GNN Layers in $\phi_\text{CVS}$ on downstream CVS performance.}
        \label{fig:cvs_gnn_layers}
    \end{subfigure}
    
    
    \caption{A series of plots showing the effect of various hyperparameters on CVS criteria prediction performance. Error bars show standard deviation.}
    \label{fig:ablation_graphs}
\end{figure*}

\noindent \textbf{Criterion-Wise Performance.} Table \ref{table:detailed_results} shows the criterion-wise performance of DeepCVS, ResNet50-DetInit, and LG-CVS; we show these results in the segmentation setting so as to better represent DeepCVS, which performs very poorly in the bounding box setting.
Introduced in Section \ref{subsec:dataset}, the three CVS criteria refer to the presence of two and only two tubular structures entering the gallbladder (C1), the clearing of fat and connective tissue from the hepatocystic triangle (C2), and the dissection of the lower part of the gallbladder from the cystic plate (C3).
Compared to DeepCVS, our model dramatically improves the classification performance for C2 and C3, while also moderately improving C1 classification performance.
Meanwhile, compared to ResNet50-DetInit, both DeepCVS and LG-CVS perform far better for C1 classification, but for C2 and C3 classification, ResNet50-DetInit is much better than DeepCVS and close to LG-CVS for C3 in particular.
This follows from the design of each method: DeepCVS and LG-CVS use semantic information more explicitly, and therefore perform very well for C1, which primarily relies on detection of the cystic duct and artery, which our detectors perform quite well for.
Meanwhile, C2 and C3 require reasoning about the level of dissection of the hepatocystic triangle and appareance of the cystic plate; our detectors more commonly fail for these classes, and therefore proper classification requires using implicit visual information to account for these errors. Consequently, ResNet50-DetInit and LG-CVS are far more effective for these criteria than DeepCVS.
In addition, our graph-based approach allows reasoning about the overall anatomical configuration in a scene, which can further improve robustness to detection errors.

\noindent \textbf{Impact of Object Detector.} To further demonstrate the robustness of our approach, we evaluate LG-CVS as well as the baseline approaches using a variety of object detectors (see Table \ref{tab:obj_detectors}).
Of note, CVS prediction performance does not correlate strongly with object detection/segmentation performance for LG-CVS or ResNet50-DetInit, but does for DeepCVS; this can be explained by the fact that DeepCVS is highly reliant on the object detections, while the other methods (ResNet50-DetInit and LG-CVS) better utilize the visual features learned by the underlying detector. As a result, the DeepCVS performance trends more closely match the object detection performance trends.
We also report the performance of DeepCVS and ResNet50-DetInit using DeepLabV3+, the original segmentation model used in~\cite{mascagni2021artificial}, and our method still outperforms these baselines.

\noindent \textbf{Qualitative Analysis.} Figure \ref{fig:qual} shows the qualitative performance of each method on four example images. On the top-left, we observe that DeepCVS and ResNet50-DetInit fail in different ways, DeepCVS incorrectly predicting C3 achievement and ResNet50-DetInit failing to identify C2 achievement. The DeepCVS failure can be attributed to the inaccuracy in object detection, where the cystic plate is incorrectly detected and segmented. Meanwhile, ResNet50-DetInit lacks semantics, and therefore must rely on the visual features corresponding to the hepatocystic triangle, which comprise only a small part of the feature map.
We see a similar mode of failure in the bottom-left example: DeepCVS and LG-CVS correctly predicted all of the criteria as not achieved, owing to incomplete dissection of the area, which manifests in the size of the predicted object masks.
In the top-right example, the detection results are again somewhat poor, with the cystic artery segmented in multiple parts; as a result, DeepCVS ends up incorrectly predicting C1 as not achieved, but LG-CVS makes the correct prediction, possibly as a result of the feature propagation through the use of a graph neural network.
Finally, in the bottom-right example, we show a failure case of LG-CVS. The detections are again incorrect, missing the cystic plate, but in this case, LG-CVS is not robust to the errors. ResNet50-DetInit, on the other hand, makes the correct prediction.

\subsection{Ablation Studies}
\label{subsec:ablations}

We additionally study the effect of various parameter settings on both the latent graph encoder $\Phi_{LG}$ and downstream CVS criteria prediction.
As our method is primarily designed for CVS prediction in the bounding box setting, we conduct all of our ablations in this setting.

\begin{table}
\caption{Latent Graph Components Ablation Study in Bounding Box Setting.}
\label{table:latent_graph_components}
\centering
\begin{tabular}{cccc}
\textbf{\begin{tabular}[c]{@{}c@{}}Visual\\ Features\end{tabular}} & \textbf{\begin{tabular}[c]{@{}c@{}}Box\\ Coordinates\end{tabular}} & \textbf{\begin{tabular}[c]{@{}c@{}}Class\\ Probabilities\end{tabular}} & \textbf{\begin{tabular}[c]{@{}c@{}}CVS Criteria\\ mAP\end{tabular}} \\ \hline
\xmark & \cmark & \cmark & 59.7 \\
\cmark & \xmark & \xmark & 60.8 \\
\cmark & \cmark & \xmark & 59.4 \\
\cmark & \xmark & \cmark & 61.1 \\
\cmark & \cmark & \cmark & \textbf{63.6} \\
\end{tabular}
\end{table}

\noindent \textbf{Latent Graph Components.} The latent graph nodes and edges are composed of box coordinates, class logits, and visual features.
Table \ref{table:latent_graph_components} shows the effect of each of these components on CVS criteria prediction performance.
Several prior works with intermediate graphical representations~\cite{wang2018videos,raboh2020differentiable,khan2021spatiotemporal} only include visual features to represent nodes and edges; however, we observe an improvement of \textbf{2.8} mAP when additionally including box coordinates and class probabilities, demonstrating their importance in the latent graph.
We also observe a decrease of \textbf{3.9} mAP when only including semantic information in the graph; taken together, these results show that LG-CVS synergistically combines semantic and visual information to predict CVS as opposed to ResNet50-DetInit, which uses only visual information, and DeepCVS, which is over-reliant on semantics.

\begin{table}[]
\caption{Effect of Edge Building Steps on LG-CVS in Bounding Box Setting.}
\label{table:edge_building}
\centering
\begin{tabular}{ccc}
\textbf{Edge Proposal} & \textbf{Edge Loss} & \textbf{CVS mAP} \\ \hline
\xmark                  & \xmark            & 60.1                     \\
\cmark                 & \xmark             & 61.5                     \\
\xmark                  & \cmark            & 62.1                     \\
\cmark                 & \cmark            & 63.6                    
\end{tabular}
\end{table}

To build the graph edges, we include edge proposal and edge classification steps. We conduct an ablation to demonstrate the benefits of each of these steps (see Table \ref{table:edge_building}).


\noindent \textbf{Box Perturbation.} As described in \ref{subsec:training_process}, we apply a random perturbation parametrized by $\lambda_{\text{perturb}}$ to the box coordinates in the latent graph $G$ prior to evaluating $\phi_{\text{CVS}}$.
Figure \ref{fig:box_perturbation} illustrates the impact of the perturbation factor $\lambda_{\text{perturb}}$ (higher corresponds to stronger perturbation).
We find that setting $\lambda_{\text{perturb}} = 0.125$ yields the best performing model.

\noindent \textbf{Reconstruction Bottleneck Size.} Prior to evaluating $\phi_{\mathcal{R}}$, we bottleneck the node visual features ($\phi_{\mathcal{R}}$ ignores the edge features) to prevent trivial solutions; this step ensures that the reconstruction objective helps the model learn a more powerful latent graph representation $G$.
Figure \ref{fig:reconstruction_bottleneck_size} illustrates the impact of varying the bottleneck size $\mathcal{F}_\mathcal{R}$ on the quality of the learned representation, reflected by the downstream performance.
We find that setting $\mathcal{F}_\mathcal{R} = 64$ yields the best performing model.

\textbf{GNN Layers.} Figure \ref{fig:cvs_gnn_layers} investigates the impact of the number of GNN layers in $\phi_{\text{CVS}}$.
We find that a 2-layer GNN architecture is most effective.


\subsection{Implementation Details}
We train all models on a single 32GB Nvidia V100 GPU using the mmdetection~\cite{chen2019mmdetection} framework, resizing all images from their original resolution of $480\times 854$ to $224\times 399$. In the first stage, we finetune a COCO-pretrained object detector for 20 epochs using a batch size of 8, each respective object detector's default hyperparameters, and the default COCO learning schedule, and select a model based on validation mAP.
In the second stage, we use the selected detector and initialize an additional ResNet50 backbone with the detector weights. We then freeze the detector and finetune the remaining weights for CVS prediction, training for 20 epochs using a batch size of 32 and AdamW optimizer with learning rate 0.00001, and finally selecting the model with the highest validation CVS mAP.
For both stages, we apply RandAugment~\cite{cubuk2020randaugment} during training.
We set $N = 16$, $E = 4$, and include GraphNorm~\cite{cai2021graphnorm} and skip connections in $\phi_{\text{LG-GNN}}$ and $\phi_{\text{CVS}}$.

\section{Conclusion}
In this work, we propose to encode surgical images as anatomy-aware latent graph representations that can be used for anatomy-reliant tasks such as Critical View of Safety prediction.
Our graph representations, which encode node and edge semantics as well as visual features to retain differentiability, can be trained without segmentation masks, utilizing far less expensive bounding box annotations instead; it can also run on top of any object detector.
In our experiments, we introduce a comprehensive evaluation paradigm for CVS criteria prediction considering different ground truth data availability scenarios, evaluating multiple baseline approaches.
We demonstrate that our method comprehensively outperforms these baselines, all while using less expensive bounding box annotations.
Finally, we show that incorporating an auxiliary image reconstruction objective improves performance across methods.
We believe that this work can serve as a template for approaching anatomy-reliant surgical video analysis tasks and promote future work exploring the use of graphical representations.

\bibliographystyle{IEEEtran}
\bibliography{main}

\end{document}